\documentclass[letterpaper, 10 pt, conference]{ieeeconf}  % Comment this line out if you need a4paper

\IEEEoverridecommandlockouts                              % This command is only needed if 
\usepackage{subcaption}
\usepackage{graphics} % for pdf, bitmapped graphics files
\usepackage{graphbox} % for vertical middle-allignment
\usepackage{epsfig} % for postscript graphics files
\usepackage{mathptmx} % assumes new font selection scheme installed
\usepackage{times} % assumes new font selection scheme installed
\usepackage{amsmath} % assumes amsmath package installed
\usepackage{amssymb}  % assumes amsmath package installed
\usepackage{graphicx}
\usepackage{multirow}
\usepackage{url}
\usepackage[noadjust]{cite}

\usepackage{xcolor}

\title{\LARGE \bf
Synthesizing Traffic Datasets using Graph Neural Networks
}

\author{Daniel Rodriguez-Criado$^{1}$, Maria Chli$^{1}$, Luis J. Manso$^{1}$, and George Vogiatzis$^{2}$% <-this % stops a space
\thanks{$^{1}$ Daniel Rodriguez-Criado, Maria Chli and Luis J. Manso are with the Computer Science Dept., Aston University, Aston Triangle, Birmingham, B4 7ET, UK.
        {\tt\small {drodr19,m.chli,l.manso}@aston.ac.uk}}%
\thanks{$^{2}$ George Vogiatzis is with the Computer Science Dept. Loughborough University, Epinal Way, Loughborough LE11 3TU {\tt\small cogv@lboro.ac.uk}}%
}

\begin{document}

\maketitle
\thispagestyle{empty}
\pagestyle{empty}

%%%%%%%%%%%%%%%%%%%%%%%%%%%%%%%%%%%%%%%%%%%%%%%%%%%%%%%%%%%%%%%%%%%%%%%%%%%%%%%%
\begin{abstract}

Traffic congestion in urban areas presents significant challenges, and Intelligent Transportation Systems (ITS) have sought to address these via automated and adaptive controls. 
However, these systems often struggle to transfer simulated experiences to real-world scenarios. 
This paper introduces a novel methodology for bridging this `sim-real' gap by creating photorealistic images from 2D traffic simulations and recorded junction footage. 
We propose a novel image generation approach, integrating a Conditional Generative Adversarial Network with a Graph Neural Network (GNN) to facilitate the creation of realistic urban traffic images.
We harness GNNs' ability to process information at different levels of abstraction alongside segmented images for preserving locality data.
The presented architecture leverages the power of SPADE and Graph ATtention (GAT) network models to create images based on simulated traffic scenarios.
These images are conditioned by factors such as entity positions, colors, and time of day.
The uniqueness of our approach lies in its ability to effectively translate structured and human-readable conditions, encoded as graphs, into realistic images. 
This advancement contributes to applications requiring rich traffic image datasets, from data augmentation to urban traffic solutions. 
We further provide an application to test the model's capabilities, including generating images with manually defined positions for various entities.

\end{abstract}

%%%%%%%%%%%%%%%%%%%%%%%%%%%%%%%%%%%%%%%%%%%%%%%%%%%%%%%%%%%%%%%%%%%%%%%%%%%%%%%%
\section{INTRODUCTION} 

Traffic congestion remains a pressing issue, particularly in major cities worldwide.
This problem is exacerbated by the increasing number of vehicles, while navigable urban space remains limited.
In this context, efficient traffic management is essential in minimising travel delays, road accidents, and environmental pollution. 
Intelligent Transportation Systems (ITS) incorporate sensing and communication technologies, along with automatic control methods to enhance the safety and efficiency of transportation infrastructure \cite{Agrawal2020}. 
\par

Junctions are critical points in traffic management as they serve as shared physical spaces accessed by multiple vehicles. 
Efficient traffic light control at intersections contributes to improved traffic flow.
The effectiveness of traditional traffic lights diminishes when they fail to adapt to changing traffic patterns \cite{Agrawal2020}. 
Machine learning advancements offer a solution by utilizing algorithms that learn optimal policies from raw sensory data.
Deep Reinforcement Learning (DRL) enables learning end-to-end models that directly control traffic light signals from CCTV footage.
Such models rely on simulated data for training due to the cost and hazards of gathering real-world traffic experience. 
Consequently, they often struggle to generalize from simulation-based training to decision-making in the real world.
% This paper focuses on bridging the `sim-real' gap by developing a tool that automatically generates photorealistic images - virtually indistinguishable from real footage - from 2D traffic simulations (e.g., SUMO \cite{behrisch2011sumo}) and recorded junction footage. 
This paper focuses on bridging the `sim-real' gap by developing a tool that automatically generates photorealistic images from 2D traffic simulations (e.g., SUMO \cite{behrisch2011sumo}) and recorded junction footage. 
Garg et al. \cite{Garg2022} recently proposed a DRL traffic light agent trained on simulated crossroads in a graphics game-like environment, addressing generalization through domain randomization \cite{Tobin2017}. 
Their technique creates diverse simulated scenarios with varying illumination, perspective, and textures, enhancing model robustness and facilitating adaptation to real-world conditions. 
However, training on photographic footage eliminates the need for domain randomization, requiring less training data for the same or better performance, as the model is trained and evaluated on similar data distribution.
\par

\begin{figure}[tb!]
    \centering
    \includegraphics[width=1\columnwidth]{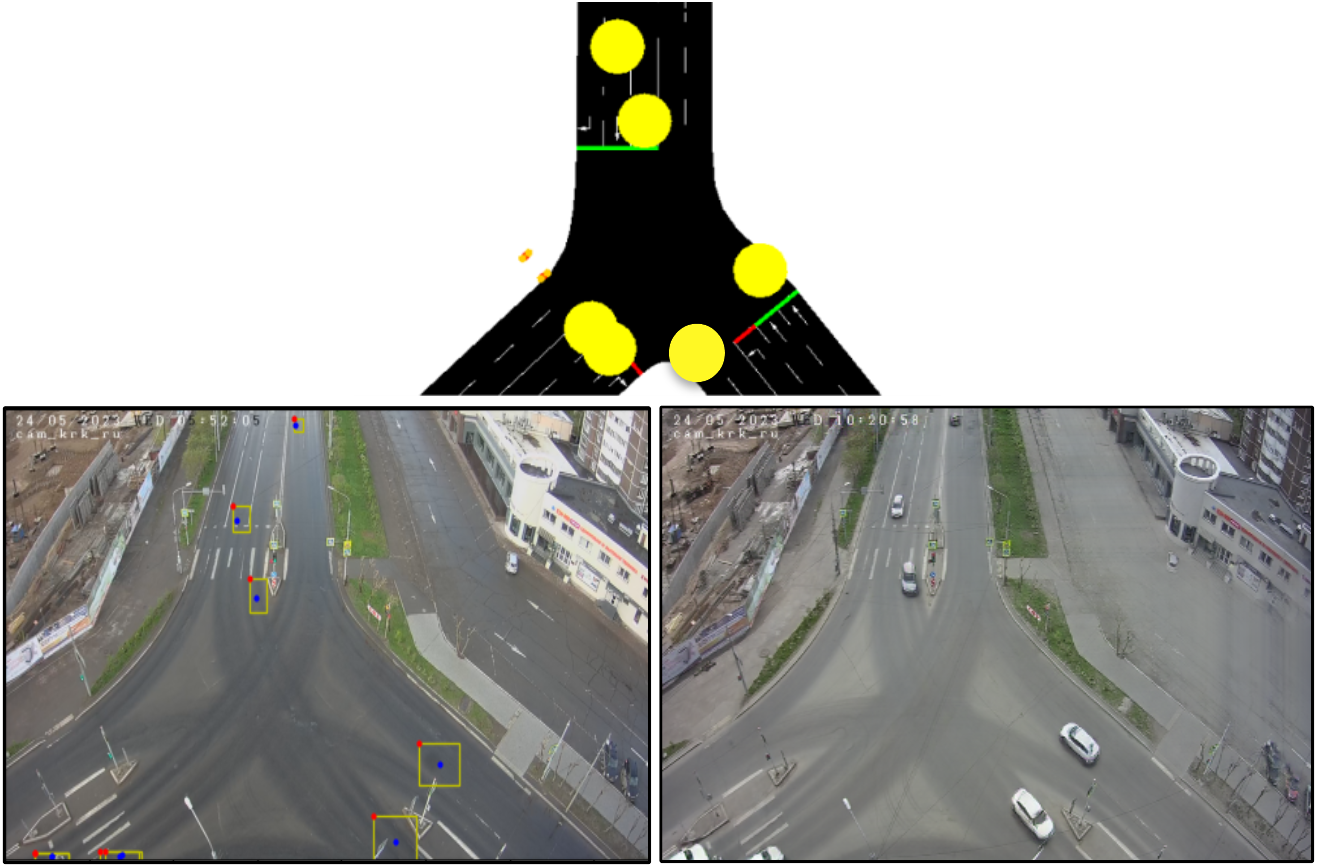}
    \caption{\textbf{The Transformation Process from SUMO-Generated to Realistic Images.} This triptych illustrates the consecutive stages involved in creating a realistic image from the SUMO simulator. The top image provides a bird's-eye view of a junction simulation in SUMO. The bottom-left image presents the corresponding bounding boxes of vehicles in SUMO, adjusted to the viewpoint of the CCTV camera. The image on the bottom right culminates the process by displaying the image generated by our model using the specified bounding boxes.}
    \vspace{-0.5cm}
    \label{fig:sumo_to_real}     
\end{figure}

This paper presents the generation of realistic urban images from simulations, focusing primarily on their application in traffic light control. 
However, the scope of this technology extends well beyond traffic management, offering a range of compelling applications. 
Consider Adaimi et al. \cite{Adaimi2020}, who utilize a drone swarm to capture aerial images of traffic for training object detection models. 
By augmenting their dataset with realistic image generation, significant improvements can be achieved in the performance and accuracy of models, as demonstrated in \cite{Barmpounakis2020}. 
Such an advancement holds immense potential for enhancing the detection and analysis of traffic patterns, thereby enabling more informed decision-making and effective urban planning. 
Traffic surveillance is another domain that stands to benefit from the generation of realistic urban images. 
Models focused on vehicle counting or tracking, utilizing camera-based systems \cite{Fernandez2021}, can increase precision and reliability through the incorporation of synthetically generated footage that encompasses diverse parameters and conditions. 
Finally, synthetically generated footage of traffic scenes under various parameters and conditions, will unlock the creation of immersive and responsive training tools for human professionals in traffic control,  empowering trainees with an enhanced understanding of the cause-and-effect relationship between their decisions and the resulting traffic dynamics, surpassing the limitations of studying historical footage alone.
\par

The persuasive impact of this technology lies in its potential to revolutionize training methodologies and decision-making processes across various domains. 
By generating realistic urban images from simulations, we can unlock unprecedented insights, refine existing models, and empower professionals in their pursuit of efficient and effective solutions.
\par

\section{BACKGROUND}
The ability to synthesize high-quality, realistic images has been a long-standing goal in computer vision.
One of the most popular applications of image generation is data augmentation \cite{arantes2020csc, mudavathu2018auxiliary}, crucial for preventing overfitting in large deep learning models. 
Others include image generation, encompassing image completion, style transfer, and resolution enhancement, among others \cite{lugmayr2022repaint, nakano2019a, enhancementReview2010}.

Three principal research directions prevail in the literature for generating realistic images: Generative Adversarial Networks (GANs)~\cite{goodfellow2020generative}, Variational AutoEncoders (VAEs)~\cite{kingma2022autoencoding}, and diffusion/denoising models~\cite{ho2020denoising}.
They are machine learning paradigms trained on real image datasets to synthesize novel images unseen in the training data.
GANs have gained considerable popularity for image generation due to their capability to produce high-resolution, diverse, and aesthetically pleasing images, as compared to the blurry images often generated by VAEs.
While denoising models also achieve high-resolution image generation, they typically demand more computational time compared to GANs and remain relatively nascent due to their recent emergence. 
For these reasons, a GAN has been selected as the image generation method for this paper.
\par

Certain GAN-based models, termed conditional GANs (cGANs), can synthesize novel images based on training data and a condition given as an image or label. 
For instance, SPADE \cite{park2019SPADE} can utilize segmentation maps as labels to generate images that appear realistic. SPADE extends the pix2pix model \cite{isola2017image}, outperforming it by better preserving semantic information in the face of typical normalisation layers (refer to Section \ref{subsecTR:SPADE} for more details).
Nonetheless, these segmentation maps only offer conditions in terms of location and object class. 
On the other hand, text-to-image synthesis models~\cite{oppenlaender2022creativity, ramesh2021zero} offer more semantic information to the generation model.
However, this comes with the trade-off of locality information, and these models are often larger as they integrate a natural language processing unit. 
To mitigate these limitations, the model proposed in this paper can be conditioned with a combination of graphs and segmented images. 
Segmented images retain locality information, while graphs, when processed by a Graph Neural Network (GNN), can introduce more abstract information.
\par

GNNs are a relatively novel learning paradigm able to process and generate graphs.
They come in numerous variants, Graph ATtention networks (GAT)~\cite{velickovic2017graph} being one of the most effective.
Graphs are of special interest because the input to many problems can intuitively be represented using them, including metric and semantic data, as well as relationships.
In urban scenarios, these data can include the time of day, weather conditions, vehicle colors, and more.
GNNs have also been proven to work well in combination with other models, such as Convolutional Neural Networks (CNNs), for image generation from graphs, as seen in the creation of cost maps for autonomous robots~\cite{rodriguez2021generation}.
\par

%Ours is a similar approach, offering a solution for generating new images to create a dataset of realistic-looking images from simulated counterparts. 
%This would establish an urban behaviour simulator from images, capable of producing synthetic footage essential for training autonomous systems in environments where data is typically sparse.

Our method generates realistic traffic intersection images based on an input graph containing the positions and colors of various entities, such as cars, trucks, buses, and pedestrians, as well as the time of day.
It is crucial to utilize a method such as GNNs for synthesizing urban scenes, as they can effectively handle a variable number of entities. 
Graphs can encode more intricate semantic information than segmented images, while segmented images convey positional information of entities. 
Once the model is trained, a traffic simulator such as SUMO \cite{behrisch2011sumo} can generate new scenarios that are subsequently easily translatable into realistic images by the proposed model.
\par

The \textbf{primary contribution} of this paper is a novel approach to image generation that integrates a cGAN model (SPADE) with a GNN to generate realistic traffic images using graphs, allowing for structured and human-readable conditioning.
To our knowledge, this is the first architecture of its kind. 
This model can transform simulated traffic crossroad scenarios into realistic images, enabling the generation of rich datasets with relative ease and minimal cost.
The resulting datasets can then serve to train various machine learning algorithms for a plethora of urban traffic applications.
Additionally, we have developed an application to generate images with vehicles and pedestrians in manually defined positions and to test the model.
More details about this tool can be found in Section~\ref{subsecTR:tool}. For comprehensive information about the entire project, please follow the URL: \url{https://vangiel.github.io/projects/traffic.html}.
\par

% Historically, traffic light control operates on fixed time frames based on historic traffic metrics \cite{koonce2008traffic}, which fail to respond to unpredictable conditions and result in time wastage for less congested roads. 
% More recently, research has focused on Adaptive Traffic Light Control (ATLC) systems, where traffic light timings are dynamically controlled based on real-time sensor data \cite{Jacome2019}.
% Numerous projects worldwide, such as SCOOT \cite{robertson1991optimizing} and ACS Lite \cite{shelby2008overview}, have adopted ATLC systems.
% These models typically rely on handcrafted signal plans from sensor data. 
% However, with numerous factors influencing traffic light control, designing an optimal signal plan manually from raw sensor data is challenging. 
% Furthermore, commonly used sensors (such as proximity sensors for vehicle counting) only provide partial information on traffic conditions, rendering the signals unable to perceive and react to real-time traffic pattern changes.\todo{This paragraph and the next are about autonomous traffic control at junctions. Try to make them more concise, focusing on how this work links to your topic.}

\section{METHOD}
This section outlines the stages involved in our method.
The first stage consists of collecting data and using such data to train a generative model (Section \ref{subsecTR:model}).
Section \ref{subsecTR:dataset}, provides a comprehensive explanation of the three components comprising the model's input: the graph, the real image, and a segmented map.
After completing the training process, the next step involves the conversion of the outputs from the SUMO simulator into graphs, as detailed in Section~\ref{subsecTR:sumo2graphs}.
These graphs are then input to our image generator model, enabling the direct generation of realistic images based on simulated inputs.
\par

Our proposed approach represents a significant advancement in the generation of realistic images by effectively leveraging the information encoded in graphs derived from simulations. 
The rest of this section delves deeper into each stage of the pipeline.
The associated code is publicly accessible at \url{https://github.com/gvogiatzis/trafficgen}.
\par

\subsection{Dataset Creation} \label{subsecTR:dataset}
\begin{figure}[htb!]
  \centering
  \begin{subfigure}[hb]{1\columnwidth}
    \centering
    \frame{\includegraphics[width=0.9\columnwidth]{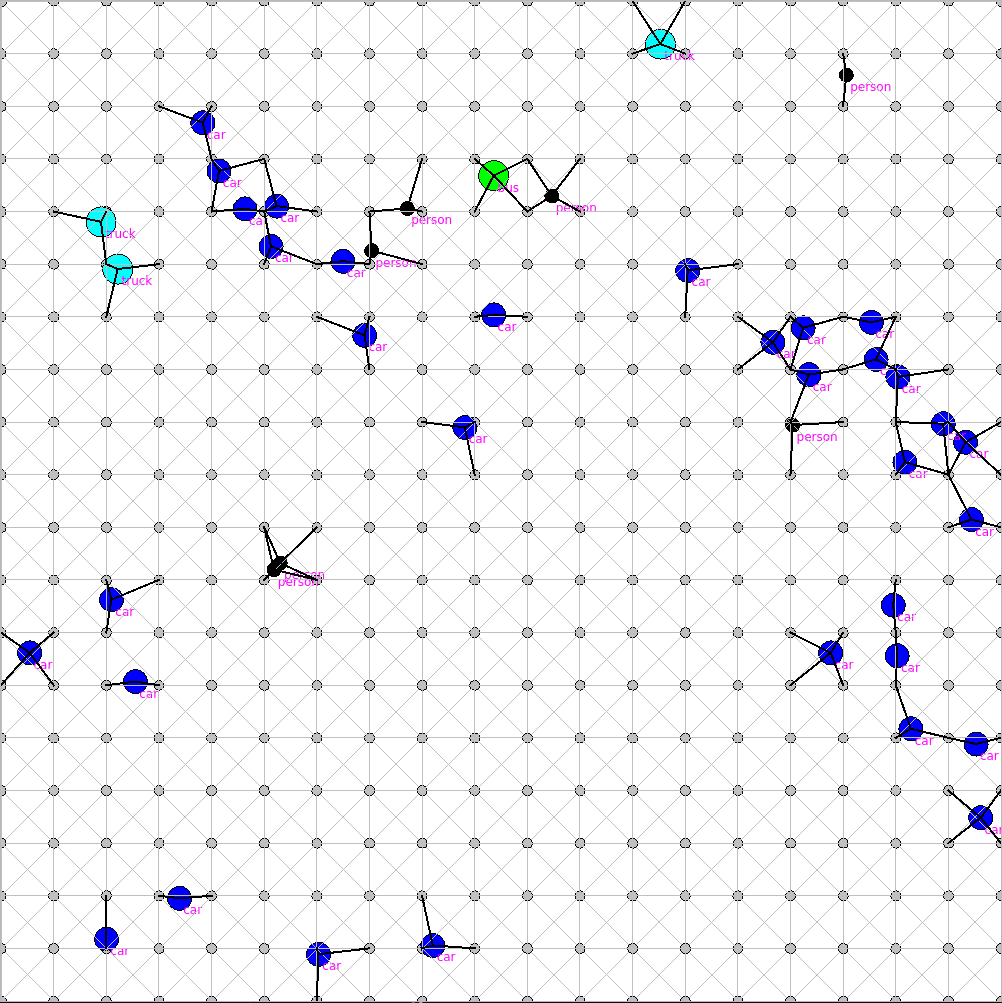}}
    \caption{\textbf{Illustration of an Input Graph.} Nodes in light gray represent grid nodes, while those in dark blue signify cars. Green nodes correspond to buses, light blue nodes denote trucks, and small black nodes represent pedestrians. The grid in this image has a resolution of $20\times20$ nodes.}
    \label{fig:input_graph_traffic}
  \end{subfigure}
  \par\bigskip
  \begin{subfigure}[b]{0.49\columnwidth}
    \centering
    \frame{\includegraphics[width=\columnwidth]{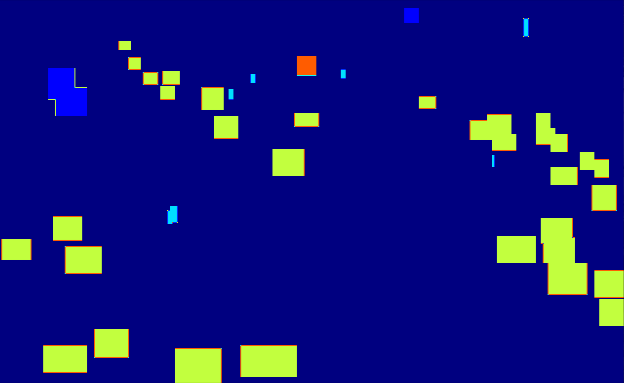}}
    \caption{\textbf{Labelled Segmented Image.} This illustration depicts the positions and sizes of the classes identified by \textit{YOLOv7} in the real image, distinguished by varying colors.}
    \label{fig:segmented_image_traffic}
  \end{subfigure}
  \begin{subfigure}[b]{0.49\columnwidth}
    \centering
    \frame{\includegraphics[width=\columnwidth]{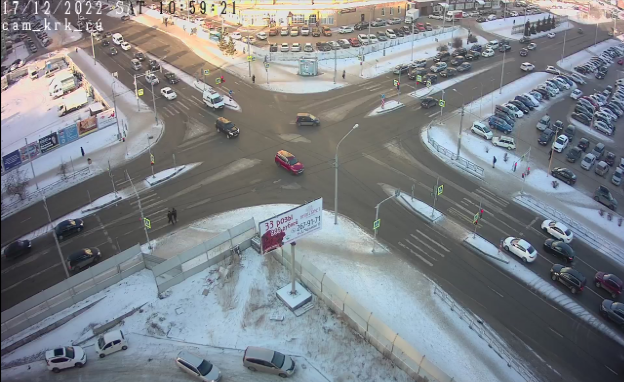}}
    \caption{\textbf{Real Image.} This image was captured from a CCTV camera positioned at a crossroad in Krasnoyarsk, Russia.}
    \vspace{0.7cm}
    \label{fig:real_image_traffic}
  \end{subfigure}
  \caption{\textbf{Illustration of a Dataset Data Point from CR1, Comprising Three Elements.} This figure incorporates the three constituent elements of the dataset: the input graph, segmented image and real image.}
  \vspace{-0.5cm}
  \label{fig:dataset_traffic}  
\end{figure}

The proposed model for image generation takes a three-element tuple as input, consisting of the segmentation map, the graph, and the real image. 
Fig. \ref{fig:dataset_traffic}  illustrates a datapoint containing these elements, providing a visual representation of the data used in this study.
This section explains the process of extracting segmentation maps and graphs from real images. 
The real images utilized for model training in this paper are sourced from two open-access real-time traffic surveillance cameras for two different traffic junctions. 
Both are situated in the city of Krasnoyarsk, Russia.
The crossroad depicted in Fig. \ref{fig:real_image_traffic} will be referred to as \textbf{CR1}\footnote{Video streaming URL of the CCTV used: \url{http://krkvideo14.orionnet.online/cam1560/embed.html?autoplay=true}}, while the one presented in Fig. \ref{fig:sumo_to_real} will be \textbf{ CR2}\footnote{Video streaming URL of the CCTV used: \url{http://krkvideo5.orionnet.online/cam1487/embed.html?autoplay=true}}.
Note that generating images for a different junction requires the collection of new data, specific to that particular setting, followed by retraining the model to accommodate the new input.
In total, $10,952$ images were collected from \textbf{ CR1} videos and $11,173$ images from \textbf{CR2} videos, in a period of 24 hours.
\par

\subsubsection{Segmentation map generation}
A segmentation map in this context is a type of image that represents different segments or regions of the original image, with each segment corresponding to a specific class present in the image.
Each pixel in a segmentation map is assigned a label that identifies the category of its corresponding pixel in the original image. 
In a visual representation of a segmentation map, each unique label is represented by a unique color, making it easy to visually distinguish between different segments or regions of the image.
\par

Generating the segmentation map is a straightforward process.
We apply an object detection model, namely \textit{YOLOv7}~\cite{wang2022yolov7}, to the actual image, yielding bounding boxes for several classes: cars, buses, trucks, and pedestrians. 
It should be noted that any other object detection model could serve the same purpose.
Importantly, the number of classes can be effortlessly extended within the range of classes detectable by \textit{YOLOv7}, if required.
\par

Utilising the normalized coordinates and dimensions of the bounding boxes for each detected object, a segmented map can be created in which each pixel value corresponds to an integer representing the class as a one-hot encoding. 
If the detection of two objects overlaps, the class from the latest detection is allocated to the pixels within the intersection—a design choice that may be modified in future work.
As the example model generates five distinct classes (including the image background), the pixel values are om the range $[0,4]$.
Fig. \ref{fig:segmented_image_traffic} shows an example of a segmented image drawing the classes with different colors.
\par

\setlength{\tabcolsep}{2pt}
\begin{table}[hbt!]
\centering
\caption{Representation of the feature vector structure for each node in the graph.}
\label{tbl:nodefeatures_traffic}
\resizebox{\columnwidth}{!}{%
\begin{tabular}{|cccccccclccc|}
\hline
\multicolumn{12}{|c|}{\textbf{Node feature vector $h$ (dimension 31 or 19)}} \\ \hline
\multicolumn{4}{|c|}{\textbf{Boxes}} & \multicolumn{5}{c|}{\textbf{Classes}}           & \multicolumn{2}{c|}{\textbf{Time}} & \textbf{Color encoding} \\ \hline
x & y & w & \multicolumn{1}{c|}{h}   & bus & truck & car & person & \multicolumn{1}{l|}{grid} & sin     & \multicolumn{1}{c|}{cos}     & \begin{tabular}[c]{@{}c@{}}clusters-colors (dimension =20)/\\ dircrete-colors (dimension=8)\end{tabular} \\ \hline
\end{tabular}%
}
\end{table}

\setlength{\tabcolsep}{3pt}

\begin{figure*}[tb!]
    \centering
    \includegraphics[width=.9\textwidth]{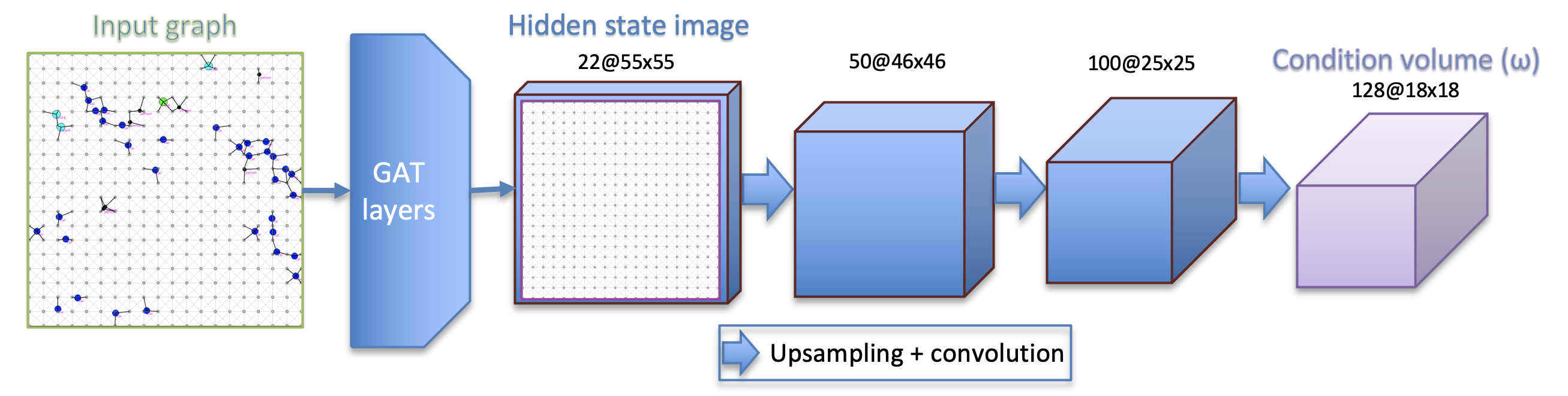}
    \caption{\textbf{Model for Creating Condition Volumes via a Combination of GNN and CNN Layers.}  This illustration shows the process of generating the condition volume $\pmb{\omega}$, which is used to condition the image generation of SPADE. The graph is processed by 3 GAT layers, and then the lattice of nodes is filtered from the output graph forming a hidden state image. This image is subsequently fed into an array of upsampling and convolutional layers to fabricate the final condition volume. Each volume depicted in the image has its dimensions written at the top, where the first number denotes the channels, and the succeeding pair specifies the height and width.}
    \vspace{-0.5cm}
    \label{fig:volumeGeneration}     
\end{figure*}

\begin{figure}[htbp]
    \centering
    \includegraphics[width=.5\textwidth]{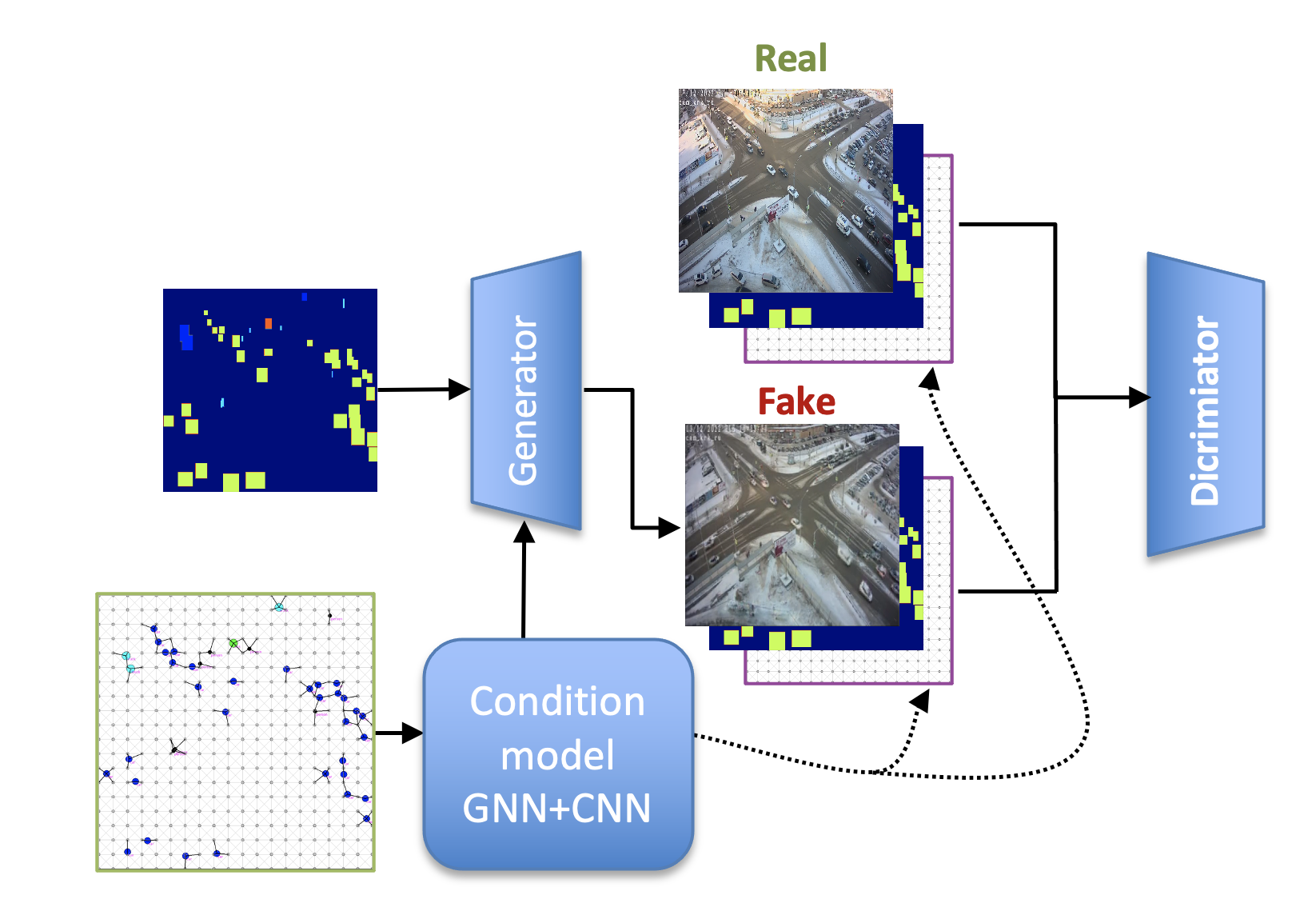}
    \caption{\textbf{Complete Model Pipeline.} The schematic depicted herein illustrates the integration of the condition model within the SPADE architecture. The condition model accepts the graph as input, producing the condition volume utilized by the generator. Moreover, the hidden state image stemming from the condition model is concatenated in the channel dimension to the discriminator's input.}
    \vspace{-0.5cm}
    \label{fig:finalModel}     
\end{figure}

\subsubsection{Graph generation}
Conversely, the creation of graphs is a critical and intricate step that offers extensive customisation flexibility to the designer. 
Due to the wide range of potential graphs that can be employed for this application, numerous variants have been explored.
For brevity this section will only discuss the two design strategies yielding the most favourable results, which only differ in how objects' colors are represented.
The first design, referred to as the \textit{clustering-colors} graph, produces the best outcomes in terms of quality of the generated image. 
On the other hand, the second design, named \textit{discrete-colors} graph, while resulting in slightly inferior performance, enables the user to condition the vehicle colors using a discrete color palette. 
This will be examined in greater detail in Section \ref{secTR:experiments}. 
Both graphs are identical, except for a minor variation in the nodes' features, as explained subsequently.
\par

The creation of the topology of the two aforementioned graph types is identical.
First, a graph is created in which each node represents an entity detected by the object detector, \textit{YOLOv7}.
A lattice of nodes is generated, with each node representing a spatial position within the image.
This grid is crucial for conditioning SPADE, as explained in Section \ref{subsecTR:volume_generation}. 
In the final step, both graphs are merged by connecting the closest entity-representing nodes to the nearest nodes in the grid within a specified radius, using the image coordinates.
Both the grid density and the connectivity radius are adjustable hyperparameters. 
Various values were assessed to achieve the optimal balance between accuracy and efficiency.
Fig. \ref{fig:input_graph_traffic} provides an example of the final graph's topology, employing a grid resolution of $20\times20$ and a connectivity radius of $1$ grid hop.
\par

As mentioned, the vectors of characteristics of the nodes differ for the two types of graphs.
The feature vector is classified into four sections, as delineated in Table \ref{tbl:nodefeatures_traffic}. The first three sections, which are shared across both graph types, encompass: the coordinates and dimensions of the object bounding box; a one-hot encoding vector of length $5$ indicating the node class (bus, truck, car, person or gird); and the encoding of the time of day using the sine and cosine.
\par

The final section of the vector, which we refer to as ``visual features'', diverges between the two graph types:
\begin{itemize}
    \item For the \textit{clustering-colors} graph: This includes a 20-element vector indicative of the object's primary color. 
This vector comprises clusters of the top $5$ most predominant colors within the bounding box, encoded in RGB.
Along with the three RGB numbers, each cluster includes an additional number with the counts normalized using a \textit{softmax} function. 
This configuration results in a total feature vector length of $31$.
    \item For the \textit{discrete-colors} graph:  Here we employ a one-hot encoding of length 8, representing the detected vehicle's color. 
The color palette includes black, white, red, lime, blue, yellow, magenta, and gray. 
Color detection consists in averaging the top $3$ most predominant colors in the bounding box and calculating the Euclidean distance to each palette color based on their RGB coordinates. 
The color with the shortest distance is chosen.
Given that averaging shifts the color closer to gray, the mean value must exceed a certain distance threshold for the specific case of selecting the gray color. 
With this option the total length of the feature vector is $19$.
\end{itemize}
\par

By integrating these features into the graph nodes, we enrich the SPADE generator with information about the entities' colors and the time of day. 
The \textit{clustering-color} graphs provide more detailed color information leading to better results, while the \textit{discrete-colors} graph enables a straightforward indication of the entity color during inference time.
The benefits of employing the \textit{discrete-colors} graph become evident when utilising the demonstration tool presented in Section \ref{subsecTR:tool}.
\par

\subsection{Model}
\label{subsecTR:model}

This section details the combination of architectures for generating the images.
The final model consists of a modification of SPADE architecture including a GNN to condition the generator with a graph.
The graph can provide richer information to the generation model allowing the creation of more complex conditions which are reflected in the final image.
First, we show the vanilla structure of SPADE alongside which we introduce our modifications.
The next subsection explains the architecture embedded in SPADE that takes graphs as inputs.
Finally, we provide a global version of the final model pipeline for generating images from graphs.
\par

\subsubsection{GAN model, SPADE}
\label{subsecTR:SPADE}
To generate images from semantic masks, SPADE layers transform segmentation masks into feature maps $\gamma$ and $\beta$ by first projecting the mask onto an embedding space that we called condition volume $\pmb{\omega}$.
Then, this volume is fed through two convolutional layers to get the feature maps. 
The generated parameters $\gamma$ and $\beta$, which are tensors with spatial dimensions, are multiplied and added to the normalized activation from the previous layer $h$, element-wise.
Thus, the activation features $h_{n,c,h,w}$ are normalized and transformed as follows:

\begin{equation}
    h'_{n,c,h,w} = \gamma_{c,h,w}(\pmb{\omega}) \frac{h_{n,c,h,w} - \mu_c}{\sigma_c} + \beta_{c,h,w}(\pmb{\omega}),
\end{equation}
where the indexes $(n, c, h, w)$ refer to the batch size, the number of channels, the height and the input width. 
The parameters $\mu_c$ and $\sigma_c$ denote the channel-wise mean and standard deviation of the input feature map $h$.
\par

The generator incorporates multiple ResNet blocks \cite{resnet} with upsampling layers.
The semantic map is downsampled to align with the resolution required for learning the modularization parameters $\gamma$ and $\beta$, as each residual block operates at a distinct scale.
In contrast, the discriminator does not employ SPADE and follows the pix2pixHD \cite{wang2018high} discriminator approach, based on PatchGAN \cite{isola2017image}, which inputs the concatenated segmentation map and the input image.
\par

The main modification introduced in this paper for the generator consists of generating the condition volume $\pmb{\omega}$ using a combination of a GNN and transpose convolutional layers as explained in Section \ref{subsecTR:volume_generation}.
Now, instead of downsampling the semantic map, it is $\pmb{\omega}$ that is rescaled to match the needed resolution in each layer.
With regards to the discriminator, the only change is the additional concatenation in the channels' dimension of information coming from the GNN to make the GAN symmetric (see Section \ref{secTR:finalModel}).
\par

\begin{figure}[tb!]
    \centering
    \includegraphics[width=0.44\columnwidth]{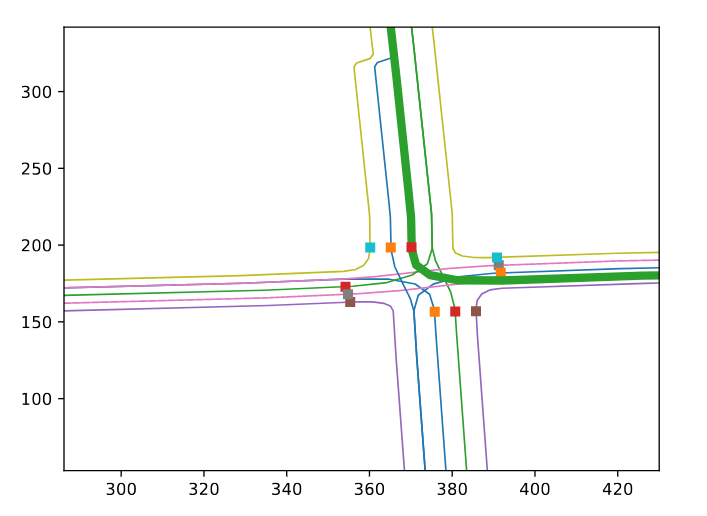}
    \includegraphics[width=0.54\columnwidth]{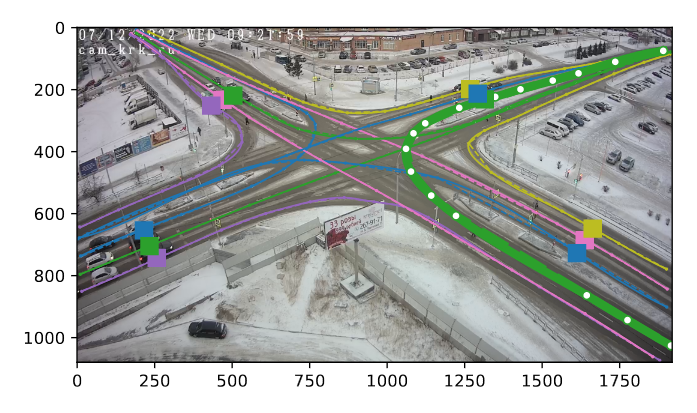}
    \caption{\textbf{The Lane Editor Application.} This illustration exhibits the interface of the lane designer application, juxtaposing the SUMO lane selector (left) with the cubic spline lane editor (right). It facilitates the definition of corresponding waypoints between the SUMO environment and real-world images for each lane spline.}
    \vspace{-0.5cm}
    \label{fig:lane_editor}     
\end{figure}

\subsubsection{Condition model}
\label{subsecTR:volume_generation}

As already mentioned, the input data (\textit{i.e.}, vehicles, pedestrians and other contextual information such as the time) are combined with a 2-dimensional lattice to form an input graph (Fig.~\ref{fig:input_graph_traffic}).
As with most GNN layers, GAT layers output graphs with the same structure as their input graph but different node embeddings \cite{velickovic2017graph}.
The input graph is processed by $3$ GAT layers, producing a graph with the same structure but adequate to condition SPADE after being filtered and processed.
This is done by training the GAT layers to embed the entities' information into the lattice sub-graph so that the lattice can be converted into an image-like format creating a latent image.
This is done by creating a pixel for each of the nodes in the lattice and using the features of the nodes as the channels of the image.
Finally, this latent image is the input of $4$ transpose convolutional layers that generate the desired condition volume $\pmb{\omega}$.
This pipeline is depicted in Fig. \ref{fig:volumeGeneration}.

\begin{figure*}[tbh!]
    \centering
    \includegraphics[width=1.9\columnwidth]{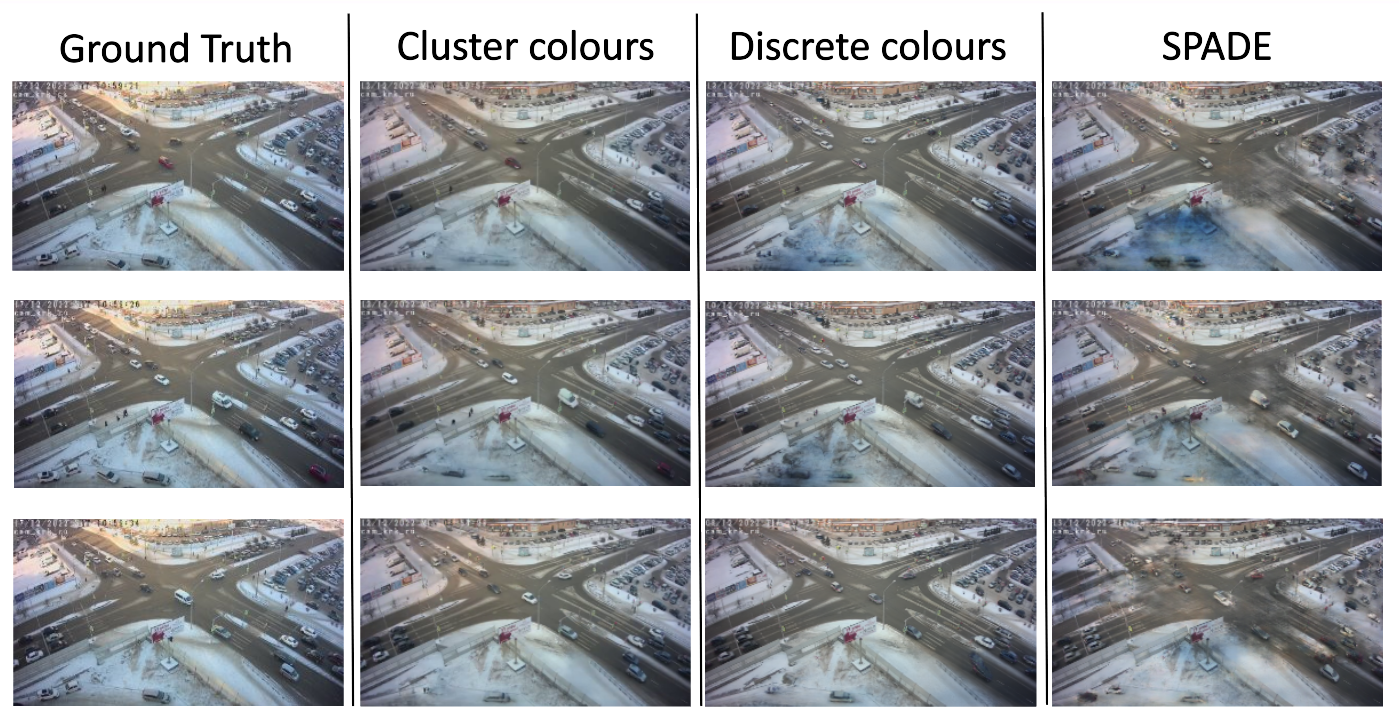}
    \caption{\textbf{Visual Results Across All Models.} This illustration presents the results for three distinctive frames from the test dataset, each in separate rows. The leftmost column constitutes the ground truth images, succeeded by images generated by the \textit{cluster-colors}, \textit{discrete-colors}, and SPADE models respectively. As observable, the \textit{cluster-colors} model is the most proficient at preserving vehicle colors, whereas the SPADE-generated images exhibit the poorest quality.}
    \vspace{-0.5cm}
    \label{fig:visual_results_traffic}     
\end{figure*}

\subsubsection{Final model}
\label{secTR:finalModel}

The final model adds the condition module to the SPADE blocks of the generator and also includes some modifications to the discriminator of SPADE to make the network symmetric.
\par

Fig. \ref{fig:finalModel} is a diagram of the final model pipeline with the generator and the discriminator of SPADE.
The data flow starts with the segmentation mask as input of the generator, then the condition module uses the graphs to yield the volumes that are also fed to the generator.
The generator produces a fake image that is concatenated in the channel dimension with the corresponding mask and the latent image of the condition module.
Finally, the real image is also concatenated with the mask and the latent image in the channel dimension, and then both the real and fake images are combined in the height dimension.
This generated fake-and-real volume is the final input of the discriminator.
Note that one fake image is generated for every two real images fed to the discriminator.
\par

\subsection{From SUMO to graphs}
\label{subsecTR:sumo2graphs}

One of the main aims of this work is to drive image synthesis from traffic simulations. One can then apply various actions to the traffic system (e.g. traffic control decisions) and observe their effect through visual footage generated in response.
This closes the loop between trial and error and enables the training of decision-making agents (AI as well as humans) from footage that is generated in response to decisions taken.
Here, we show an instance of this loop closure using SUMO \cite{behrisch2011sumo} as the driving simulator. 
\par

The first step is to create a topologically faithful representation of the traffic junction network within SUMO. 
This is readily achieved using scripts that can import geometry and network structure from OpenStreetMap (OSM) data. 
The second step involves defining the correspondence between 2D points on the SUMO simulation to points on the real-world junction. 
If the OSM data was geometrically faithful, the mapping would be a simple 2D homography (8 degrees of freedom). In practice, there are significant discrepancies between the simulation geometry and the real-world junction. 
The solution we adopt in this paper is to define individual traffic lanes in the junction as cubic splines. This is achieved in a few minutes of clicking through points in the junction images and works well in practice. Fig. \ref{fig:lane_editor} shows the lane designer GUI with the sumo lane selector (left) and the cubic spline lane editor (right).
Within each lane spline, we can define corresponding way-points between SUMO and the real world, (e.g. the point where cars stop for the red light and other well-defined landmarks). At the end of this process, we have a reliable mapping between points on each SUMO lane and the corresponding points on the real junction.
\par

The final stage of this process involves determining vehicle bounding boxes for each location within the junction. 
These bounding boxes subsequently facilitate the generation of the graphs, as described in Section \ref{subsecTR:dataset}, which serve as inputs to our image generation model. 
To obtain these bounding boxes, we fit a spatial bounding box distribution using histograms.
Fig. \ref{fig:sumo_to_real} illustrates the pipeline from a SUMO frame (left) to a set of bounding boxes defined on the real junction image (middle) and the synthesized CCTV frame (right) containing the road background and vehicles in the right locations. 
\par

\section{EXPERIMENTATION AND RESULTS}
\label{secTR:experiments}

\begin{table*}[tb!]
\centering
\caption{Results for the three different models evaluated in the proposed metrics.}
\label{tab:metrics_by_class}
\resizebox{\textwidth}{!}{%
\begin{tabular}{|c|l|ccc|ccc|}
\hline
\multirow{2}{*}{\textbf{Models}} & \multicolumn{1}{c|}{\multirow{2}{*}{\textbf{FID}}} & \multicolumn{3}{c|}{\textbf{mIoU}}                             & \multicolumn{3}{c|}{\textbf{Accu.}}                             \\ \cline{3-8} 
                                 & \multicolumn{1}{c|}{}                              & Cars                & People             & Trucks              & Cars                & People              & Trucks              \\ \hline
\textbf{SPADE}                   & 176.32835                                          & \textbf{0.63349355} & \textbf{0.3243129} & 0.099982            & \textbf{0.70595584} & \textbf{0.59278189} & 0.13323193          \\
\textbf{cluster-colors}         & \textbf{149.88987}                                 & 0.53274998          & 0.21035392         & \textbf{0.27845053} & 0.69294361          & 0.24580686          & \textbf{0.34741428} \\
\textbf{discrete-colors}        & 154.56592                                          & 0.50205496          & 0.16974847         & 0.01920121          & 0.66411157          & 0.20146478          & 0.02427287          \\ \hline
\end{tabular}%
}
\vspace{-0.5cm}
\end{table*}

\subsection{Implementation details}
 
All experiments were carried out utilizing an NVIDIA RTX A6000 GPU that has a memory capacity of 48GB. 
All the models were trained with image resolutions set at 640x640 pixels, whereas YOLOv7 was used to detect bounding boxes from images with a resolution of 1280x1280 pixels. 
A batch size of 12 was employed for the training phase, and this was increased to 24 during testing.

\subsection{Dataset and metrics}

We trained three different models with the \textbf{CR1} dataset:  the standard SPADE version, and the combination of the GNN and SPADE for the two types of graphs.
These models were trained using an identical training set consisting of $10,322$ images, graphs, and segmentation maps. 
Each model was evaluated using three distinct metrics on a test set comprising $630$ data points, following the same metric system utilized by SPADE \cite{park2019SPADE}.
To quantify segmentation accuracy, we used mean Intersection-over-Union (mIoU) and pixel accuracy (Accu.).
The Fr\`echet Inception Distance (FID) \cite{heusel2017gans}, on the other hand, was utilized to evaluate the discrepancy between the distributions of synthetic and real images.
\par

We calculated the mIoU and pixel accuracy metrics for each class, excluding the `background' due to its disproportionate size relative to other classes. 
The `buses' category was also omitted due to a lack of sufficient images within the training dataset to facilitate a decent generation of these vehicles.
The performance against these metrics is summarized in Table \ref{tab:metrics_by_class}.
\par
% \begin{table}[htb!]
% \centering
% \caption{}
% \label{tbl:general_metrics}
% \resizebox{\columnwidth}{!}{%
% \begin{tabular}{|c|c|c|c|}
% \hline
% Models           & mIoU               & Accu.              & FID                \\ \hline
% SPADE            & 0.9100028921329103 & 0.9528169681609624 & 176.32835964151167 \\ \hline
% cluster-colors  & 0.8768255680705048 & 0.9342816607762896 & 149.8898718540088  \\ \hline
% discrete-colors & 0.8660801365675568 & 0.9281635742187501 & 154.56592412507683 \\ \hline
% \end{tabular}%
% }
% \end{table}

Our analysis of the results reveals that our models demonstrate a significant enhancement in the FID score compared to SPADE, with the \textit{cluster-colors} model producing superior outcomes for this metric. 
Although the mIoU and pixel accuracy of our models are generally slightly lower than SPADE results, they remain competitive, with the \textit{cluster-colors} model producing superior results for trucks. 
Bear in mind that the FID is the most critical measure for this study, as it evaluates the realism of the generated images, whereas the other two metrics pertain more directly to image segmentation models.
Nevertheless, we have included the mIoU and pixel accuracy as these metrics were employed in the original SPADE paper.
\par

It is also worth highlighting that the additional computational burden of the \textit{cluster-colors} and \textit{discrete-colors} models over the basic SPADE model is marginal, adding only $10.42\%$ and $10.28\%$ more parameters, respectively. 
Consequently, the computational speed and memory usage remain comparable.
The training for our models and SPADE for the specified batch size and number of images took approximately $3.5$ days. 
The generation of images for the test set takes approximately $2$ minutes for all the models excluding the time needed to load the data.
\par

% \subsection{Parameters}
% SPADE:
% Number of generator parameters: 92233557
% Number of discriminator parameters: 92233557
% Number of total parameters: 184467114

% Clusters:
% Number of generator parameters: 101840769
% Number of discriminator parameters: 101840769
% Number of total parameters: 203681538

% Colors:
% Number of generator parameters: 101711169
% Number of discriminator parameters: 101711169
% Number of total parameters: 203422338

\subsection{Visual results}

In order to interpret the above metrics more tangibly, we present several frames generated by each model in Fig. \ref{fig:visual_results_traffic}.
Each row in this figure represents a different frame from the dataset, where the first image corresponds to the ground truth derived from the test set, and the subsequent images are the outputs generated by the \textit{cluster-colors}, \textit{discrete-colors}, and \textit{vanilla} SPADE models, in sequence.
\par

A close inspection of the results reveals that the \textit{cluster-colors} model is proficient at effectively reconstructing the colors of the majority of vehicles present in the original images. 
However, the \textit{discrete-colors} model tends to struggle in generating vibrant colors, although it performs adequately with white, black, and gray vehicles. 
This issue could potentially be attributed to the method employed to discretize the color palette, a point that could be optimized in future work. 
In contrast, images synthesized by the vanilla SPADE model display vehicles with arbitrary colors, and they often introduce artefacts into the majority of frames, rendering them less visually appealing and realistic.
\par

\subsection{Time conditioning}

\begin{figure}[tbh!]
    \centering
    \includegraphics[width=1\columnwidth]{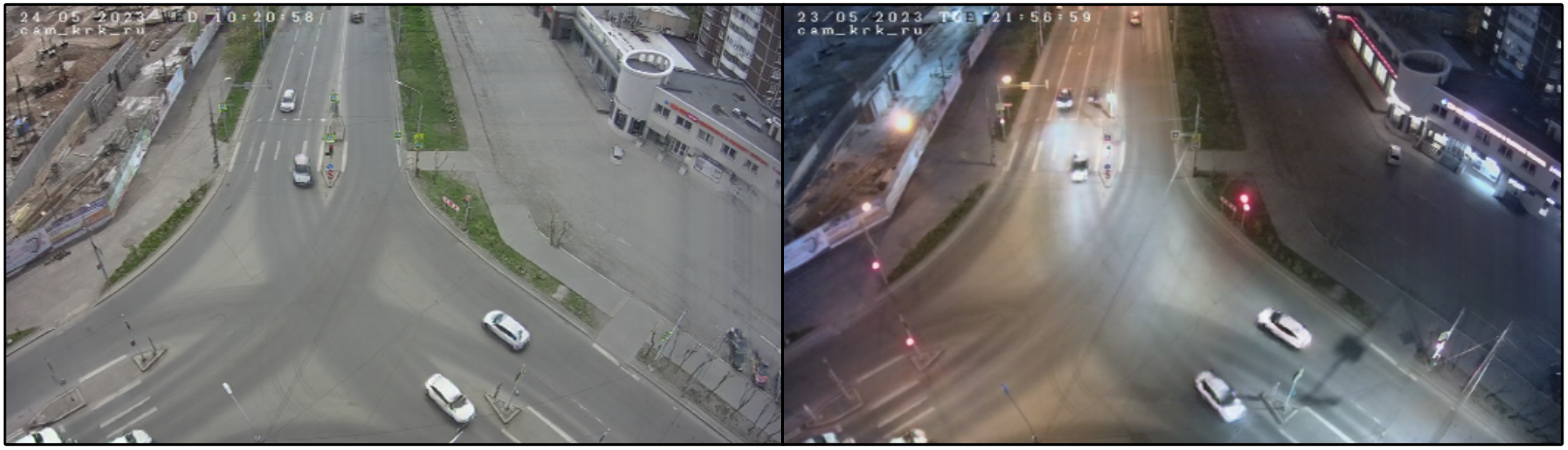}
    \caption{\textbf{Two Examples for Images Generated in Different Times of the Day.} These images were generated using the demo tool in Section \ref{subsecTR:tool}. It is readily apparent that the daytime-generated image (left) possesses more illumination, whereas the nocturnal counterpart (right) exhibits a darker ambience, authentically simulating the respective time frames.}
    \vspace{-0.5cm}
    \label{fig:time_conditioning}     
\end{figure}

As previously discussed, the proposed model in this study is also capable of generating images conditioned at different times of the day. Fig. \ref{fig:time_conditioning} demonstrates this capability, featuring two images generated by the \textit{discrete-colors} model corresponding to daytime (Fig. \ref{fig:time_conditioning} left) and nighttime conditions (Fig. \ref{fig:time_conditioning} right).

\subsection{Interactive tool}
\label{subsecTR:tool}

\begin{figure}[tbh!]
    \centering
    \includegraphics[width=1\columnwidth]{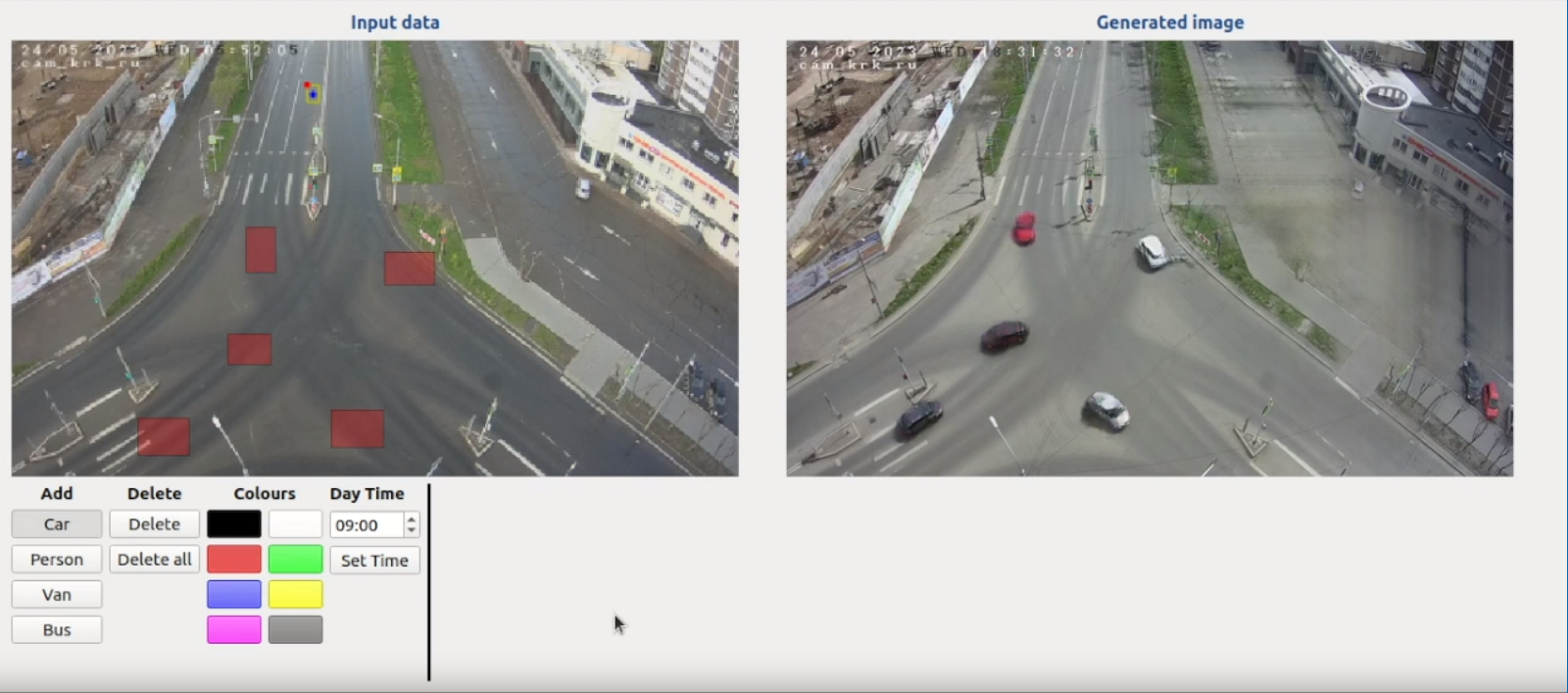}
    \caption{\textbf{Graphical User Interface of the Interactive Tool.} The application's interface is divided into two principal frames: the left frame facilitates the entry of user input data, while the right frame displays the generated image. The toolbox for choosing vehicle types, colors, and time of day – parameters to condition the image generation – is located in the lower-left corner.
    % Users can sketch an arbitrary number of bounding boxes within the left frame, indicating the desired positions of entities to be generated.
    }
    \vspace{-0.1cm}
    \label{fig:demoTool}     
\end{figure}

To evaluate the image generation model under various conditions, a tool with a visual interface was designed, as illustrated in Fig. \ref{fig:demoTool}. 
The graphical user interface (GUI) displays two distinct frames.
On the left frame, the user can draw various bounding boxes, each denoting the placement of entities to be generated.
\par

The entity type, its color, and the time of day can be specified using the tools button, situated at the lower-left corner of the GUI. 
The right frame exhibits the image produced by the model, reflecting the entities and conditions specified within the left frame.
A new image is generated every time a modification is done in the left frame.
\par

\section{CONCLUSIONS}
\label{secTR:conclusions}

The synthesis of realistic images from simulated ones is highly beneficial and presents a myriad of applications, where data augmentation stands out.
In the present paper, we have accomplished the development of a tool that is capable of generating traffic images with a realistic appearance from a simulator by merging a GAN-based model, SPADE, with a GNN for conditioning.
\par

This developed tool enables the production of large datasets with relatively less effort, vital for training computationally demanding deep learning models that comprize numerous parameters (among other applications), thereby preventing the overfitting of the training data. 
This results in enhanced generalisation to new scenarios. 
Section \ref{secTR:experiments} illustrates the effectiveness of our model in comparison to the unmodified version of SPADE, with only a minor increase in computational complexity.
\par

Our model is not only capable of generating realistic images but also conditioning features of the generated images using semantic information, namely the colors of the vehicles in the image and the time of the day.
Section \ref{subsecTR:tool} shows a tool with a GUI that enables the user to produce images by manually setting the stated conditions along with the positions of the entities to be created.
\par

Looking towards future research, we plan to train the model with additional data and utilize more than one consecutive frame as input for the model. 
These improvements are likely to enhance the quality of the generated images.
We also aim to expand our experimentation by including more classes for the model to generate and adding more conditions such as the weather.
An intriguing prospect for subsequent research involves substituting the current cGAN with a diffusion model.
Given the noteworthy results achieved by these models in recent years, this approach could yield valuable insights.
\par

\addtolength{\textheight}{-12cm}   % This command serves to balance the column lengths

%%%%%%%%%%%%%%%%%%%%%%%%%%%%%%%%%%%%%%%%%%%%%%%%%%%%%%%%%%%%%%%%%%%%%%%%%%%%%%%%

%%%%%%%%%%%%%%%%%%%%%%%%%%%%%%%%%%%%%%%%%%%%%%%%%%%%%%%%%%%%%%%%%%%%%%%%%%%%%%%%

%%%%%%%%%%%%%%%%%%%%%%%%%%%%%%%%%%%%%%%%%%%%%%%%%%%%%%%%%%%%%%%%%%%%%%%%%%%%%%%%

%\listoftodos

%\section*{APPENDIX}


%\section*{ACKNOWLEDGMENT}

%The preferred spelling of the word ÒacknowledgmentÓ in America is without an ÒeÓ after the ÒgÓ. Avoid the stilted expression, ÒOne of us (R. B. G.) thanks . . .Ó  Instead, try ÒR. B. G. thanksÓ. Put sponsor acknowledgments in the unnumbered footnote on the first page.

%%%%%%%%%%%%%%%%%%%%%%%%%%%%%%%%%%%%%%%%%%%%%%%%%%%%%%%%%%%%%%%%%%%%%%%%%%%%%%%%

%References are important to the reader; therefore, each citation must be complete and correct. If at all possible, references should be commonly available publications.

\bibliographystyle{IEEEtran}
\bibliography{mybibfile}

\end{document}